\documentclass{article}

\usepackage[final,nonatbib]{Neurips_2021}

\usepackage[utf8]{inputenc} 
\usepackage[T1]{fontenc}    
\usepackage{hyperref}       
\usepackage{url}            
\usepackage{booktabs}       
\usepackage{amsfonts}       
\usepackage{nicefrac}       
\usepackage{microtype}      
\usepackage{xcolor}         
\usepackage{graphicx}
\usepackage{subcaption}
\usepackage{float}
\usepackage{multirow}
\usepackage{listings}
\usepackage{caption} 
\usepackage{upquote}

\newcommand\YAMLcolonstyle{\color{red}\mdseries}
\newcommand\YAMLkeystyle{\color{black}\bfseries}
\newcommand\YAMLvaluestyle{\color{blue}\mdseries}

\makeatletter
\newcommand\language@yaml{yaml}

\expandafter\expandafter\expandafter\lstdefinelanguage
\expandafter{\language@yaml}
{
  keywords={true,false,null,y,n},
  keywordstyle=\color{darkgray}\bfseries,
  basicstyle=\YAMLkeystyle,                                 
  sensitive=false,
  comment=[l]{\#},
  morecomment=[s]{/*}{*/},
  commentstyle=\color{purple}\ttfamily,
  stringstyle=\YAMLvaluestyle\ttfamily,
  moredelim=[l][\color{orange}]{\&},
  moredelim=[l][\color{magenta}]{*},
  moredelim=**[il][\YAMLcolonstyle{:}\YAMLvaluestyle]{:},   
  morestring=[b]',
  morestring=[b]",
  literate =    {---}{{\ProcessThreeDashes}}3
                {>}{{\textcolor{red}\textgreater}}1     
                {|}{{\textcolor{red}\textbar}}1 
                {\ -\ }{{\mdseries\ -\ }}3,
}

\lst@AddToHook{EveryLine}{\ifx\lst@language\language@yaml\YAMLkeystyle\fi}
\makeatother

\newcommand\ProcessThreeDashes{\llap{\color{cyan}\mdseries-{-}-}}

\title{Semantic Augmentation in Images using Language}

\author{Sahiti Yerramilli*  \And Jayant Sravan Tamarapalli*
    \And Tanmay Girish Kulkarni*  \And Jonathan Francis \quad Eric Nyberg \\ \\
    Carnegie Mellon University \\ \newline
    \texttt{\{syerrami,jtamarap,tgkulkar\}@alumni.cmu.edu \quad \{jmf1,en09\}@andrew.cmu.edu }\\
}

\begin{document}

\maketitle

\begin{abstract}
Deep Learning models are incredibly data-hungry and require very large labeled datasets for supervised learning. As a consequence, these models often suffer from overfitting, limiting their ability to generalize to real-world examples. Recent advancements in diffusion models have enabled the generation of photorealistic images based on textual inputs. Leveraging the substantial datasets used to train these diffusion models, we propose a technique to utilize generated images to augment existing datasets. This paper explores various strategies for effective data augmentation to improve the out-of-domain generalization capabilities of deep learning models.
\end{abstract}

\section{Introduction}
Supervised deep learning has traditionally thrived on the availability of extensive labeled datasets, enabling models to learn patterns and make accurate predictions. However, as the field progresses and larger data-hungry models, such as Vision Transformers \cite{DBLP:journals/corr/abs-2010-11929} \cite{dai2021coatnet}, emerge, the challenge of finding datasets that scale accordingly becomes increasingly daunting. This scarcity of labeled data poses a significant obstacle to the development and training of these advanced models.

Interestingly, while labeled datasets for visual tasks might be limited, there exists an abundance of large text corpora that have been utilized to train large transformer models \cite{raffel2020exploring} \cite{lan2020albert} \cite{he2021deberta}. These transformer models have exhibited remarkable performance across a diverse range of natural language processing tasks, demonstrating their capability to effectively learn and represent complex linguistic patterns.

Recently, text-conditioned image generation models have made significant progress in terms of the diversity and the photorealism of the generated images \cite{DBLP:journals/corr/abs-2102-12092} \cite{DBLP:journals/corr/abs-2112-10741} \cite{DBLP:journals/corr/abs-2112-10752}. These models, commonly referred to as diffusion models, have demonstrated the ability to generate photorealistic images based on textual inputs. This breakthrough opens up new avenues for data augmentation in the realm of computer vision.

The primary objective of this paper is to delve into various augmentation strategies based on the diffusion models suitable for the image classification task. We conduct experiments using the COCO Captions dataset \cite{chen2015microsoft} and introduce modifications to the captions based on four distinct strategies: prefix, suffix, replacement, and compound. Detailed explanations of these strategies will be provided in the method \ref{method} section. Additionally, we briefly explore how a model trained with these augmentations can be employed for other computer vision-based classification tasks.

Our paper first goes into the related work associated with the different components used within our approach. This is followed by the approach, experiment design, and experimental results. Finally, we summarize our results in the conclusion and provide additional training details and cluster setup within the appendix.

\section{Related work}
    
    \subsection{Generative Models based Augmentation}
    Several previous works have explored different GAN-based approaches to address the challenge of data scarcity and augmentation. For example, studies such as \cite{https://doi.org/10.48550/arxiv.1711.04340}, have utilized CycleGAN to enrich domains with fewer examples by transforming images from domains with a higher abundance of examples. Similarly, \cite{Huang_2018_ECCV} employed Generative Adversarial Networks (GANs) to generate new images, which aligns with our approach.

    However, there are notable limitations in these existing works. In the case of \cite{https://doi.org/10.48550/arxiv.1711.04340}, the focus primarily lies on domain adaptation, requiring the learning of a new CycleGAN for each domain shift. Moreover, \cite{https://doi.org/10.48550/arxiv.1912.02781} does not perform well for multi-label classification, as it lacks the ability to learn relationships between labels and can only handle one class at a time. Additionally, these approaches lack disentanglement in the latent space, making targeted modifications to specific parts of an image challenging. Furthermore, they do not adequately consider the contextual information within the images.
    
    Another relevant work is DatasetGAN \cite{DBLP:journals/corr/abs-2104-06490}, which leverages the latent space disentanglement of StyleGAN \cite{DBLP:journals/corr/abs-1812-04948}. This model incorporates a style interpreter attached to the StyleGAN model, enabling the production of pixel-wise segmentation annotations with minimal human annotations. However, a drawback of this approach is that StyleGAN is typically trained to generate a single class of objects, necessitating the use of multiple StyleGAN models for augmentation across different classes. Furthermore, this method also faces limitations in augmenting scenarios involving multiple labels.
    
    \subsection{Image Transformation Based Augmentation}
    
    Studies such as \cite{https://doi.org/10.48550/arxiv.1912.02781}, \cite{https://doi.org/10.48550/arxiv.1710.09412}, and \cite{https://doi.org/10.48550/arxiv.1909.13719} have proposed different strategies for augmenting datasets by enhancing existing images through various transformations.

    In \cite{https://doi.org/10.48550/arxiv.1912.02781}, multiple instances of transformations are applied to the same image, effectively augmenting the dataset. Similarly, \cite{https://doi.org/10.48550/arxiv.1710.09412} employs a linear combination of different images from the dataset, where the coefficient of the linear combination is sampled from a beta distribution. These approaches demonstrate efforts to augment the dataset; however, a limitation arises in that the resulting augmentation is still a combination of existing dataset images and may not produce meaningful or realistic images.
    
    Furthermore, the focus of these methods primarily lies in improving in-domain performance, without thoroughly discussing the potential benefits of out-of-domain generalization. While these techniques contribute to data augmentation techniques, there remains a need to explore approaches that generate diverse and meaningful images while also considering their potential for enhancing model performance in real-world scenarios beyond the training dataset.
        
\section{Semantic Augmentation} \label{method}
\begin{figure}[H]
        \centering
            \includegraphics[width=0.7\linewidth]{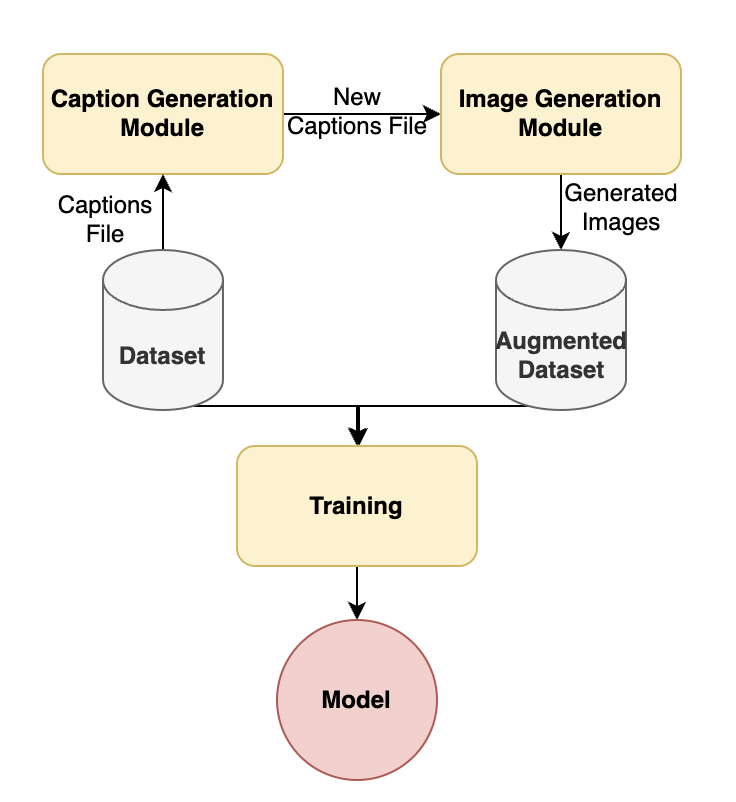}
          \caption{Semantic Augmentation Pipeline}
          \label{fig:overall-pipeline}
\end{figure}

The core concept of our approach involves modifying the captions associated with images in the dataset and leveraging these newly generated captions to create corresponding images using a text-to-image diffusion model, specifically the Stable Diffusion model \cite{DBLP:journals/corr/abs-2112-10752}. The main pipeline comprises two key components: the caption generation module and the image generation module. \ref{fig:overall-pipeline} presents a comprehensive overview of the entire pipeline, illustrating the interconnection and interaction between the various modules involved in the process. In the subsequent sections, we provide a detailed exploration of each module, unraveling the intricacies of the end-to-end system we have developed. For our experiments, we leverage the COCO Captions dataset \cite{chen2015microsoft}, which provides a collection of images, each accompanied by a set of captions.

\subsection{Caption Generation}
\subsubsection{Caption Label Extraction}
\begin{figure}[H]
        \centering
            \includegraphics[width=1\linewidth]{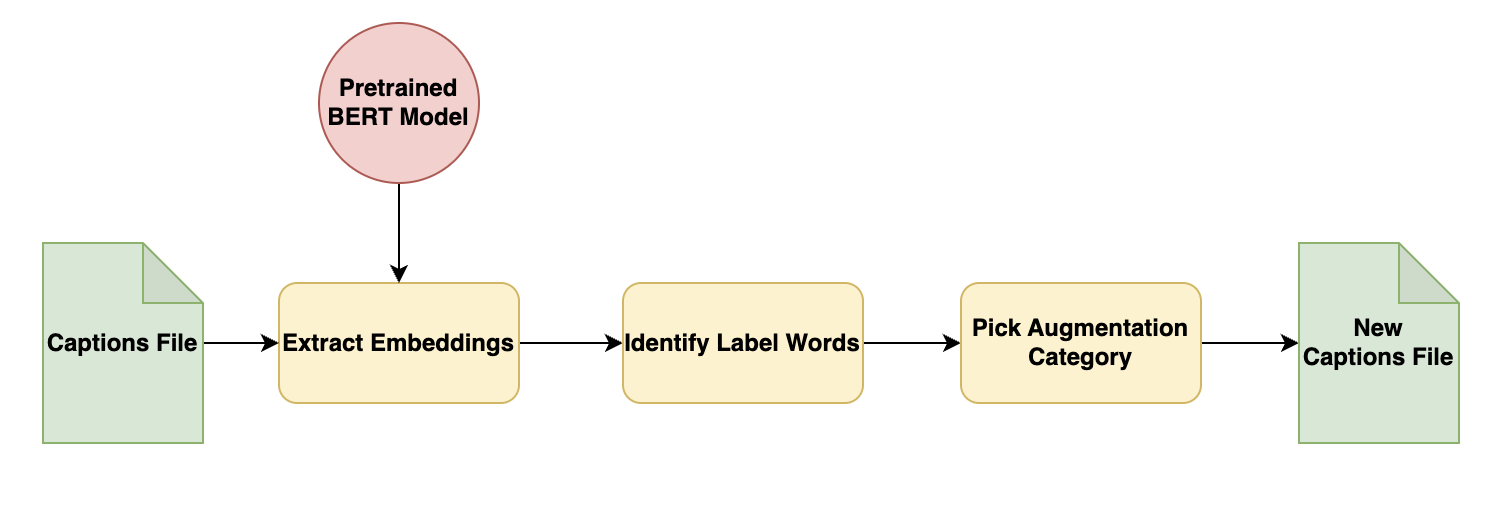}
          \caption{New Caption Generation Process}
          \label{fig:caption-generation}
\end{figure}

For a comprehensive overview of the  caption generation process, refer to \ref{fig:caption-generation}. It provides a visual representation of the different stages and steps involved in generating captions for the given task. While the categories in the COCO Captions dataset are predefined, the specific word forms used in the captions may vary. For instance, if an image depicts a "woman sitting on a couch," the corresponding label word in the caption should be "woman," which belongs to the "person" class. To modify or change labels within the captions, it is essential to identify the closest word in the caption that corresponds to the desired class. To accomplish this, we can employ pre-trained language models trained on extensive text corpora. In our methodology, we utilize the BERT model \cite{DBLP:journals/corr/abs-1810-04805} to extract embeddings for all the words in the caption. By calculating the cosine distance between the embedding of each word and the embedding of the class label, we can identify the closest word to the class label within the caption. By performing label extraction and finding the closest words within the captions, we establish a mechanism to modify or replace labels in the augmentation process. This step ensures that the generated images are associated with appropriate labels that align with the desired class definitions.
\subsubsection{Augmentation Methods}
We employ four distinct augmentation methods for each caption in our approach:
\begin{enumerate}
    \item \textbf{Prefix Augmentation}: This method involves appending a predefined prefix to the original caption. The prefix is selected from a predetermined list, which includes variations such as "A cartoon of," "A grainy image of," or "A black and white image of." By adding these prefixes, we introduce additional contextual information to the caption.

    \item \textbf{Suffix Augmentation}: Similar to prefix augmentation, this method appends a predefined suffix to the original caption. The suffix is chosen from a predefined list, which includes variations such as "on a rainy day," "on a foggy night," "in the mountains," or "near the sea." By incorporating these suffixes, we enhance the description of the image with additional details.

    \item \textbf{Replacement Augmentation}: In this method, we focus on the labels within the COCO Captions dataset. Each label belongs to a specific supercategory (e.g., animal, person, vehicle). We randomly sample a subset of labels within an image and replace them with other labels from the same supercategory. This replacement ensures that the augmented caption maintains coherence and remains relevant to the image content.

    \item \textbf{Compound Augmentation}: This method combines all the previously defined augmentation techniques in a sequential manner. By applying the prefix, suffix, and replacement augmentations successively, we create a compound augmentation that introduces a comprehensive range of modifications to the original caption.
\end{enumerate}
\begin{figure}[H]
        \centering
            \includegraphics[width=1\linewidth]{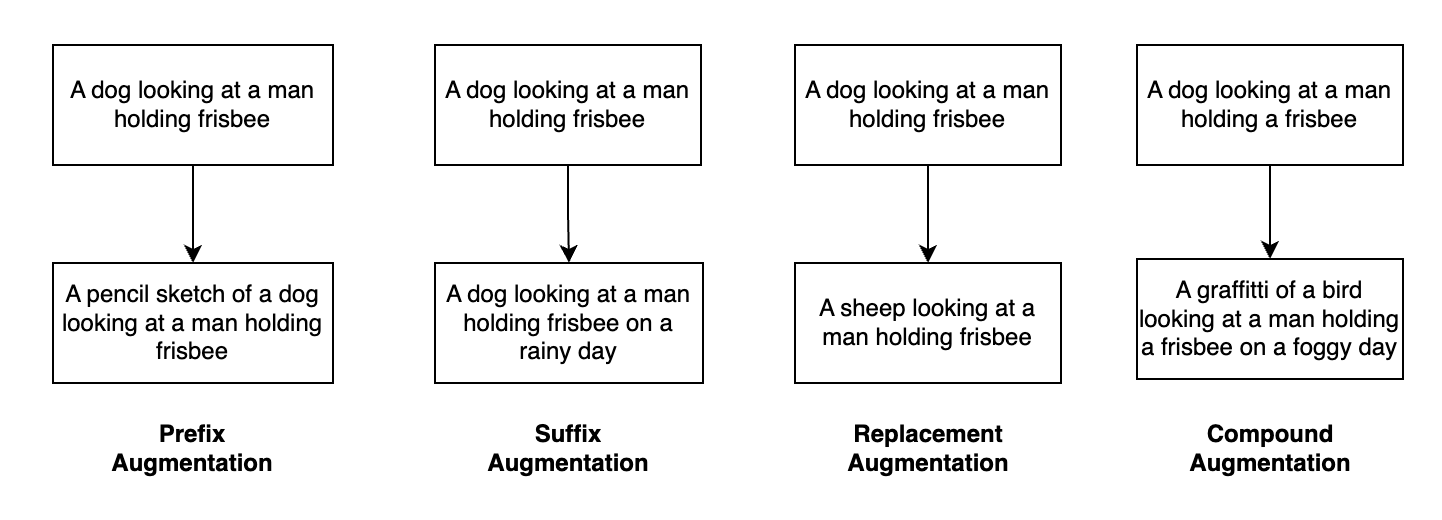}
          \caption{Examples from Each Augmentation Category}
          \label{fig:aug-ex}
\end{figure}
For every caption in the dataset, we randomly select an augmentation strategy and apply it to generate a new caption. \ref{fig:aug-ex} gives an example for each augmentation category.

\subsection{Image Generation}
Next, after obtaining the set of newly generated captions through augmentation, our focus shifts to generating photorealistic images that correspond to these modified captions. To achieve this, we employ a diffusion model, specifically the Stable Diffusion model \cite{DBLP:journals/corr/abs-2112-10752}. This choice is based on the model's ability to generate a diverse range of classes effectively.
\begin{figure}[H]
        \centering
            \includegraphics[width=1\linewidth]{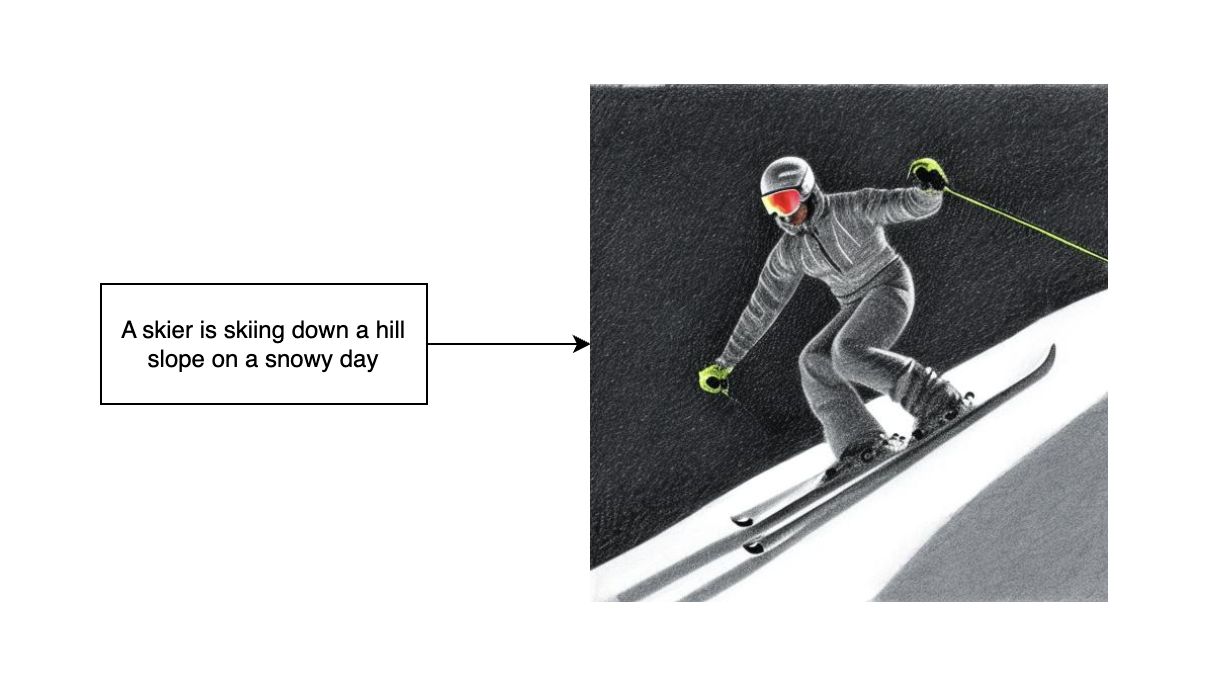}
          \caption{Text-to-Image Generation Example}
          \label{fig:image-gen}
\end{figure}
By leveraging the capabilities of the Stable Diffusion model, we can produce high-quality images that align with the augmented captions. It allows us to generate images that encompass various objects, scenes, and visual concepts. \ref{fig:image-gen} shows an example of such a generation.

To facilitate convenient usage and compatibility with existing tools and frameworks, we store the generated images along with their respective labels in the COCO Dataset format. This format standardizes the representation of images and associated annotations, simplifying their integration into subsequent stages of the pipeline.

\subsection{Augmentation}
Lastly, utilizing the generated images, we proceed to augment the original COCO Captions dataset during the training of our classification models. The augmentation process involves incorporating the generated images alongside the existing dataset.

An important hyperparameter to consider is the number of augmented images to be included per original image in the dataset. Fine-tuning this hyperparameter is crucial to achieve optimal performance. By systematically tuning this parameter, we aim to strike a balance between incorporating enough augmented images to enrich the dataset and avoiding overrepresentation or bias that may impact model performance.

\section{Experiments}
\subsection{Datasets}
We conducted our experiments using the COCO Captions dataset \cite{chen2015microsoft}, which comprises a large collection of images accompanied by their respective captions. This dataset serves as the primary source of training and evaluation for our classification model. Additionally, we utilized other relevant datasets, such as the PASCAL VOC \cite{10.1007/s11263-009-0275-4} and Tiny Imagenet \cite{Le2015TinyIV}, for conducting comparative analyses and assessing the generalization capabilities of our models.

\subsection{In-Domain Performance}

In this subsection, we present the results of our experiments that evaluate the performance of the classification model within the same domain as the training dataset (COCO). We measure the performance of this model using well-established metrics like mean average precision(mAP) and accuracy.

Furthermore, we compare our models against existing state-of-the-art generalization techniques - Mixup \cite{https://doi.org/10.48550/arxiv.1710.09412} and AugMix \cite{https://doi.org/10.48550/arxiv.1912.02781} to demonstrate the advancements. Refer to table \ref{tab: in-domain} for the results.

\begin{table}[H]
            \begin{center}
            \begin{tabular}{|c|c|c|c|}
                \hline
                \textbf{Model} & \textbf{mAP} & \textbf{Accuracy}  \\
                \hline
                Vanilla Resnet & 0.529 & 0.974 \\
                \hline
                Augmix & 0.558 & 0.975 \\
                \hline
                Mixup & 0.552 & 0.972 \\   
                \hline
                Semantic Augmentation & \textbf{0.564} & \textbf{0.975}\\
                \hline 
            \end{tabular}\\
        \caption{\label{tab: in-domain} In-Domain Performance on COCO}
        \end{center}
    \end{table}

\subsection{Out-of-domain Performance}

To assess the generalization capabilities of our models, we conducted experiments on PASCAL VOC \cite{10.1007/s11263-009-0275-4}. For each dataset, we followed a transfer learning approach. Specifically, we replaced the last layer of our pre-trained classification model with a new layer corresponding to the number of classes in the respective dataset. We froze all other layers to preserve the learned features and only trained the newly added layer. This allowed us to evaluate the performance of our models on the specific task of classifying images within these out-of-domain datasets.

By conducting these out-of-domain experiments, we aim to assess the transferability of our models and their potential to generalize to diverse visual datasets. The results of these experiments shed light on the robustness of our models and their ability to handle out-of-domain challenges.

\begin{table}[H]
            \begin{center}
            \begin{tabular}{|c|c|c|c|}
                \hline
                \textbf{Model} & \textbf{mAP} & \textbf{Accuracy}  \\
                \hline
                Vanilla Resnet & 0.652 & 0.952 \\
                \hline
                Augmix & 0.675  & 0.954 \\
                \hline
                Mixup & 0.672 & 0.952 \\   
                \hline
                Semantic Augmentation & \textbf{0.702} & \textbf{0.957}\\
                \hline 
            \end{tabular}\\
        \caption{\label{tab: out-domain2} Out-of-Domain Performance on PASCAL VOC}
        \end{center}
    \end{table}

\subsection{Analysis}
The analysis of the experimental results highlights the superior performance of our augmentation model in both in-domain and out-of-domain experiments. Our model consistently outperformed all other models considered in the evaluation. Future research should focus on conducting extensive out-of-domain experiments on various datasets to validate the generalization capabilities of our model more comprehensively. Another potential avenue is to explore fine-tuning techniques for the Stable Diffusion model used in image generation. Fine-tuning the diffusion model can enable it to better capture the nuances and complexities of different visual domains, ultimately improving the generalization capability of the augmented images it generates.

\nocite{jain2023maea}
\bibliographystyle{plain}
\bibliography{ref.bib}

\begin{thebibliography}{10}

\bibitem{https://doi.org/10.48550/arxiv.1711.04340}
Antreas Antoniou, Amos Storkey, and Harrison Edwards.
\newblock Data augmentation generative adversarial networks, 2017.

\bibitem{chen2015microsoft}
Xinlei Chen, Hao Fang, Tsung-Yi Lin, Ramakrishna Vedantam, Saurabh Gupta, Piotr Dollar, and C.~Lawrence Zitnick.
\newblock Microsoft coco captions: Data collection and evaluation server, 2015.

\bibitem{https://doi.org/10.48550/arxiv.1909.13719}
Ekin~D. Cubuk, Barret Zoph, Jonathon Shlens, and Quoc~V. Le.
\newblock Randaugment: Practical automated data augmentation with a reduced search space, 2019.

\bibitem{dai2021coatnet}
Zihang Dai, Hanxiao Liu, Quoc~V. Le, and Mingxing Tan.
\newblock Coatnet: Marrying convolution and attention for all data sizes, 2021.

\bibitem{DBLP:journals/corr/abs-1810-04805}
Jacob Devlin, Ming{-}Wei Chang, Kenton Lee, and Kristina Toutanova.
\newblock {BERT:} pre-training of deep bidirectional transformers for language understanding.
\newblock {\em CoRR}, abs/1810.04805, 2018.

\bibitem{DBLP:journals/corr/abs-2010-11929}
Alexey Dosovitskiy, Lucas Beyer, Alexander Kolesnikov, Dirk Weissenborn, Xiaohua Zhai, Thomas Unterthiner, Mostafa Dehghani, Matthias Minderer, Georg Heigold, Sylvain Gelly, Jakob Uszkoreit, and Neil Houlsby.
\newblock An image is worth 16x16 words: Transformers for image recognition at scale.
\newblock {\em CoRR}, abs/2010.11929, 2020.

\bibitem{10.1007/s11263-009-0275-4}
Mark Everingham, Luc Gool, Christopher~K. Williams, John Winn, and Andrew Zisserman.
\newblock The pascal visual object classes (voc) challenge.
\newblock {\em Int. J. Comput. Vision}, 88(2):303–338, jun 2010.

\bibitem{he2021deberta}
Pengcheng He, Xiaodong Liu, Jianfeng Gao, and Weizhu Chen.
\newblock Deberta: Decoding-enhanced bert with disentangled attention, 2021.

\bibitem{https://doi.org/10.48550/arxiv.1912.02781}
Dan Hendrycks, Norman Mu, Ekin~D. Cubuk, Barret Zoph, Justin Gilmer, and Balaji Lakshminarayanan.
\newblock Augmix: A simple data processing method to improve robustness and uncertainty, 2019.

\bibitem{Huang_2018_ECCV}
Sheng-Wei Huang, Che-Tsung Lin, Shu-Ping Chen, Yen-Yi Wu, Po-Hao Hsu, and Shang-Hong Lai.
\newblock Auggan: Cross domain adaptation with gan-based data augmentation.
\newblock In {\em Proceedings of the European Conference on Computer Vision (ECCV)}, September 2018.

\bibitem{jain2023maea}
Vidhi Jain, Jayant~Sravan Tamarapalli, Sahiti Yerramilli, and Yonatan Bisk.
\newblock Maea: Multimodal attribution for embodied ai.
\newblock {\em arXiv preprint arXiv:2307.13850}, 2023.

\bibitem{DBLP:journals/corr/abs-1812-04948}
Tero Karras, Samuli Laine, and Timo Aila.
\newblock A style-based generator architecture for generative adversarial networks.
\newblock {\em CoRR}, abs/1812.04948, 2018.

\bibitem{lan2020albert}
Zhenzhong Lan, Mingda Chen, Sebastian Goodman, Kevin Gimpel, Piyush Sharma, and Radu Soricut.
\newblock Albert: A lite bert for self-supervised learning of language representations, 2020.

\bibitem{Le2015TinyIV}
Ya~Le and Xuan~S. Yang.
\newblock Tiny imagenet visual recognition challenge.
\newblock 2015.

\bibitem{DBLP:journals/corr/abs-2112-10741}
Alex Nichol, Prafulla Dhariwal, Aditya Ramesh, Pranav Shyam, Pamela Mishkin, Bob McGrew, Ilya Sutskever, and Mark Chen.
\newblock {GLIDE:} towards photorealistic image generation and editing with text-guided diffusion models.
\newblock {\em CoRR}, abs/2112.10741, 2021.

\bibitem{raffel2020exploring}
Colin Raffel, Noam Shazeer, Adam Roberts, Katherine Lee, Sharan Narang, Michael Matena, Yanqi Zhou, Wei Li, and Peter~J. Liu.
\newblock Exploring the limits of transfer learning with a unified text-to-text transformer, 2020.

\bibitem{DBLP:journals/corr/abs-2102-12092}
Aditya Ramesh, Mikhail Pavlov, Gabriel Goh, Scott Gray, Chelsea Voss, Alec Radford, Mark Chen, and Ilya Sutskever.
\newblock Zero-shot text-to-image generation.
\newblock {\em CoRR}, abs/2102.12092, 2021.

\bibitem{DBLP:journals/corr/abs-2112-10752}
Robin Rombach, Andreas Blattmann, Dominik Lorenz, Patrick Esser, and Bj{\"{o}}rn Ommer.
\newblock High-resolution image synthesis with latent diffusion models.
\newblock {\em CoRR}, abs/2112.10752, 2021.

\bibitem{https://doi.org/10.48550/arxiv.1710.09412}
Hongyi Zhang, Moustapha Cisse, Yann~N. Dauphin, and David Lopez-Paz.
\newblock mixup: Beyond empirical risk minimization, 2017.

\bibitem{DBLP:journals/corr/abs-2104-06490}
Yuxuan Zhang, Huan Ling, Jun Gao, Kangxue Yin, Jean{-}Francois Lafleche, Adela Barriuso, Antonio Torralba, and Sanja Fidler.
\newblock Datasetgan: Efficient labeled data factory with minimal human effort.
\newblock {\em CoRR}, abs/2104.06490, 2021.

\end{thebibliography}
    
\end{document}